\documentclass[letterpaper,journal]{IEEEtran}
\usepackage{amsmath,amsfonts}
\usepackage{xcolor}
\usepackage{array}
\usepackage[caption=false,font=normalsize,labelfont=sf,textfont=sf]{subfig}
\usepackage{textcomp}
\usepackage{stfloats}
\usepackage{url}
\usepackage{verbatim}
\usepackage{graphicx}
\usepackage{cite}
\usepackage{mathtools}
\usepackage{algorithm}
\usepackage{algpseudocode}
\usepackage{amssymb}
\usepackage{hyperref}
\usepackage[capitalise,nameinlink]{cleveref}
\usepackage{booktabs} 
\usepackage{tabularx}
\usepackage{amsthm}

\begin{document}

\title{CG-FKAN: Compressed-Grid Federated Kolmogorov–Arnold Networks for Communication Constrained Environment

\author{Seunghun Yu, Youngjoon Lee, Jinu Gong \IEEEmembership{Member, IEEE}, Joonhyuk Kang, \IEEEmembership{Member, IEEE}}

\thanks{This research was partly supported by the Institute of Information \& Communications Technology Planning \& Evaluation (IITP)-ITRC (Information Technology Research Center) grant funded by the Korea government (MSIT) (IITP-2025-RS-2020-II201787, contribution rate: 50\%) and (IITP-2025-RS-2023-00259991, contribution rate: 25\%). Also, this research was financially supported by Hansung University for Jinu Gong (contribution rate: 25\%). \\(\textit{Corresponding author: Joonhyuk Kang})
}

\thanks{S. Yu, Y. Lee and J. Kang are with the School of Electrical Engineering, KAIST, Daejeon, South Korea. \\(e-mails: \{sh0703.yu, yjlee22, jkang\}@kaist.ac.kr)
}

\thanks{
J. Gong is with the Department of Applied AI, Hansung University, Seoul, South Korea. (e-mail: {jinugong}@hansung.ac.kr)
}

}
\maketitle

\begin{abstract}
Federated learning (FL), widely used in privacy-critical applications, suffers from limited interpretability, whereas Kolmogorov–Arnold Networks (KAN) address this limitation via learnable spline functions.
However, existing FL studies applying KAN overlook the communication overhead introduced by grid extension, which is essential for modeling complex functions.  
In this letter, we propose CG-FKAN, which compresses extended grids by sparsifying and transmitting only essential coefficients under a communication budget. 
Experiments show that CG-FKAN achieves up to 13.6\% lower RMSE than fixed-grid KAN in communication-constrained settings. 
In addition, we derive a theoretical upper bound on its approximation error.

\end{abstract}

\section{Introduction}
\IEEEPARstart{W}{ith} the rapid development of wireless communications, large volumes of data collected from clients can now be utilized for machine learning~\cite{zhu2020toward}. 
However, directly aggregating such data to a central server raises privacy concerns~\cite{kairouz2021advances}.
Federated learning (FL) emerges as an effective paradigm that enables collaborative model training while preserving data locality and mitigating privacy risks in centralized aggregation~\cite{mcmahan2017communication,yang2021federated}.
Given this property, FL is suitable for applications involving sensitive information, such as healthcare and defense applications~\cite{9415623,khan2021federated}.

However, many AI models employed in such privacy-critical applications operate as black boxes, making it difficult to interpret or ensure the trustworthiness of their decisions~\cite{nascita2024survey}.
This limitation remains particularly critical in FL, as it is widely applied to sensitive domains where transparency and interpretability are essential~\cite{jagatheesaperumal2024enabling}.
In this context, the Kolmogorov–Arnold Network (KAN) is gaining increasing attention in FL, as its learnable spline functions have the potential to offer more transparent representations than fixed activations.
Prior studies apply KAN to various FL scenarios—including ECG classification task~\cite{Zeleke2024}, federated benchmarks~\cite{lee2025unifiedbenchmarkfederatedlearning,Zeydan2025}, and hybrid KAN–LSTM framework for traffic prediction~\cite{11106287}—exploring its promise as a viable approach to enable FL with KAN.

\begin{figure}
    \includegraphics{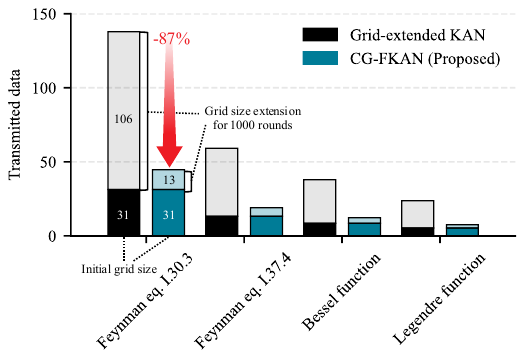} 
    \caption{
        Comparison of the required data between the grid-extended KAN and CG-FKAN in a regression task.
        The figure shows the required data transmitted by all participating clients in each round, with the bottom bar indicating the initial grid and the stacked bars showing data transmitted to the server as the grid size increases.
    }
    \label{fig:Intro_graph}
\end{figure}

Despite these advances, existing studies on FL with KAN do not examine how grid structures affect communication efficiency or expressiveness.
In practice, grid extension in KAN is crucial for accurately modeling complex nonlinear functions~\cite{liu2024kan1}.
While grid extension enhances the representational capacity of the model, it also causes a substantial increase in the number of parameters, thereby exacerbating communication overhead in FL settings.
To reduce the communication cost, we propose the Compressed-Grid Federated KAN (CG-FKAN), which mitigates communication overhead by sparsifying grid coefficients and transmitting only the most informative ones to the server.
Although various approaches are proposed to compress model parameters in FL for different architectures ~\cite{Gong2023,Yi_kong2024}, this work is the first to address the parameter explosion caused by grid extension in KAN under communication-constrained conditions.
As illustrated in~\cref{fig:Intro_graph}, proposed CG-FKAN significantly reduces the transmitted data compared to the grid-extended KAN.
The main contributions of this letter are summarized as follows:
\begin{itemize}
\item We analyze the grid-extended KAN within FL, highlighting its substantial impact on communication efficiency.
\item We propose the Compressed-Grid Federated KAN (CG-FKAN), achieving competitive performance with significantly reduced communication overhead.
\item We further derive a theoretical upper bound on the approximation error of CG-FKAN, demonstrating its proximity to the optimal solution.
\end{itemize}

\section{System Setup}\label{ch2:system}
\subsection{FL Framework}
We consider an FL system consisting of \(N\) clients and a central server.  
Each client \(k\) owns a private dataset $\mathcal{D}_{k} = \{({x}_k, y_k)\}$, where \({x}_k\) and \(y_k\) denote the input and output samples, respectively. 
Note that, the global dataset is \(\mathcal{D} = \bigcup_{k=1}^{N}\mathcal{D}_{k}\), where local data across clients follow a Dirichlet distribution to reflect the non-IID nature of FL \cite{li2022federated}.
To preserve data privacy, the server aggregates model parameters without directly accessing client data.
Under this setting, the global optimization objective in FL is defined as:
\begin{align}
\min_{w_{g}\in\mathbb{R}^d} F(w_{g})
  &\coloneqq \frac{1}{N}\sum_{k=1}^{N} f_{k}(w_{g}),\\
\text{where } f_{k}(w_{g})
  &\coloneqq \frac{1}{|\mathcal{D}_{k}|}\sum_{({x}_k,y_k)\in\mathcal{D}_k}
    \mathcal{L}(w_{g};{x}_k,y_k),
\end{align}
and \(\mathcal{L}(\cdot\,;{x}_k,y_k)\) denotes the local loss function.  
At each round \(t \in \{0,\dots,T\}\), the server selects a random subset of clients \(\mathcal{S}_t \subseteq \{1,\dots,N\}\) and broadcasts the current global model \(w_{g}^{(t-1)}\).  
Each participating client \(k \in \mathcal{S}_t\) updates its local model \(w_{k}^{(t)}\) using \(\mathcal{D}_k\) and uploads the result to the server.  
The server then aggregates the updates:  
\begin{align}
w_{g}^{(t)}
  &= \frac{1}{|\mathcal{S}_t|}\sum_{k\in\mathcal{S}_t} w_{k}^{(t)}.\label{eq:aggregate_paramters}
\end{align}

\begin{figure*}[ht]
    \centering
    \includegraphics{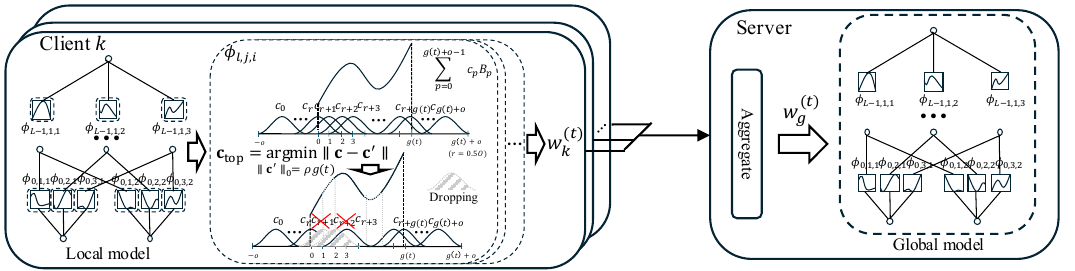} 
    \caption{Overview of CG-FKAN. 
    At round $t$, each client $k$ trains its local KAN model and performs sparsification by dropping coefficients in all spline functions.
    To this end, it retains only coefficients that minimize pairwise differences from neighboring coefficients, thereby preserving the most informative variations.
    Only the remaining coefficients are transmitted to the server as the parameter vector ${w}_k^{(t)}$.
    Then, the server aggregates the sparsified updates from clients to construct the global model ${w}_g^{(t)}$.
    }
    \label{fig:proposed_diagram}
\end{figure*}

\subsection{FL with Kolmogorov–Arnold Networks}
We extend the conventional FL framework by employing KAN as the local models at each client, while keeping the global model \(w_g\) and loss function \(\mathcal{L}(\cdot)\) unchanged.  
Each client uses a KAN with \(L\) layers, where layer \(l\) contains \(n_l\) nodes.  
Let \(\phi_{l,j,i}(\cdot)\) denote the activation function on the edge from node \(i\) in layer \(l\) to node \(j\) in layer \(l{+}1\).  
Given an input \({x}_k\), the KAN output at client \(k\) is expressed as:
\begin{equation}
h_k({x}_k)
= \sum_{i_{L-1}=1}^{n_{L-1}}
  \phi_{L-1,i_L,i_{L-1}}\Bigl(\cdots
  \sum_{i_0=1}^{n_0}\phi_{0,i_1,i_0}(x_{k,i_0})\Bigr).
\end{equation}

Each activation function can be represented by a B-spline expansion as:
\begin{equation}
\phi_{l,j,i}(x)
= \frac{\alpha x}{1 + e^{-x}}
  + \sum_{p=0}^{g+o-1} c_p B_p(x),
\end{equation}
where \(\alpha\) is a learnable coefficient, \(B_p(x)\) denotes the \(p\)-th B-spline basis, and \(c_p\) represents its corresponding weight.  
Here, \(g\) and \(o\) denote the number of grid points and the spline order, respectively.  

In FL, the local objective at round \(t\) is given by
\begin{equation}
\min_{w_g}\; f_k\bigl(w_g^{(t-1)}\bigr)
= \frac{1}{|\mathcal{D}_k|}\sum_{({x}_k,y_k)\in\mathcal{D}_k}
  \mathcal{L}\bigl(h_k({x}_k;w_g^{(t-1)}), y_k\bigr),
\end{equation}
and each client updates its model parameters as:
\begin{equation}
w_k^{(t)} 
\leftarrow w_g^{(t-1)} 
  - \eta\, \nabla f_k\bigl(w_g^{(t-1)}\bigr),
\end{equation}
where \(\eta\) denotes the learning rate.  
Finally, the server aggregates the local updates \(\{w_k^{(t)}\}\) according to~\eqref{eq:aggregate_paramters}.

\subsection{Grid Extension for KAN-based FL} 
To enhance the performance of KAN in FL, we adopt a fine-graining strategy that progressively increases the grid size at fixed training intervals~\cite{liu2024kan1}.  
Let \(g_0\), \(T_p\), and \(\{\delta_j\}_{j=1}^{\lfloor T/T_p\rfloor}\) denote the initial grid size, the grid extension period (in rounds), and the sequence of incremental grid-size increases, respectively.  
Then, at round \(t\), the grid size is defined as:  
\begin{equation}
  g(t) = g_0 + \sum_{j=1}^{\lfloor t/T_p\rfloor}\delta_j.
\end{equation}

Whenever \(t\) is an integer multiple of \(T_p\), the server broadcasts \(\bigl(w_g^{(t-1)}, g(t)\bigr)\) to all clients.  
At this point, each client expands its grid by inserting \(\delta_{\lfloor t/T_p\rfloor}\) new knots, trains the enlarged model, and returns the updated parameters to the server.  
This fine-graining process improves model performance, yet simultaneously increases the per-round communication load by \(\delta_{\lfloor t/T_p\rfloor}\) additional parameters per client.

\section{CG-FKAN}\label{ch3:proposed}
In this section, we introduce the proposed CG-FKAN, designed to reduce uplink communication cost under grid extension in KAN-based FL.  
The overall process of CG-FKAN is illustrated in~\cref{fig:proposed_diagram}.

\subsection{Compression via Sparsification}
Let \(C_0\) denote the fixed bit overhead, \(b\) the number of bits per coefficient, and \(\omega_i = n_i n_{i+1}\) the number of connections between layers \(i\) and \(i{+}1\).
Without sparsification, the required uplink communication cost is given as:
\begin{align}
    R_u(t) = b\Bigl(C_0 + \sum_{i=0}^{L-1}\omega_i\,g(t)\Bigr).
\end{align}

As grid extension increases spline coefficients over rounds, full transmission notably raises communication cost.  
Thus, we adopt a sparsification scheme that retains only key coefficients under a fixed uplink budget \(R_{\rm budget}\).  
In particular, each client selects the maximum \(\rho \in [0,1]\) that meets this constraint.
The total transmission cost \(R(\rho)\) consists of two components: the payload \(R_{\text{pay}}(\rho)\) and the position-encoding bits \(R_{\text{pos}}(\rho)\)~\cite{Sattler2019SparseBinaryCompression}, i.e., $R(\rho) = R_{\text{pay}}(\rho) + R_{\text{pos}}(\rho)$, where \(R_{\text{pay}}(\rho) = \rho \sum_i \omega_i\,g(t)\) and \(R_{\text{pos}}(\rho) = \sum_i \omega_i \log_2 \binom{g(t)+o}{\rho\,g(t)}\).  
Therefore, \(\rho\) is selected as the maximum ratio, which satisfies the communication constraint as:
\begin{align}\label{eq2:ratio}
  \rho = \max\Bigl\{r \,\Big|\,
    &b\bigl(C_0 + r\sum_{i=0}^{L-1}\omega_i\,g(t)\bigr)\notag\\
    &+ \sum_{i=0}^{L-1}\omega_i\log_2\binom{g(t)+o}{r\,g(t)}
    \le R_{\rm budget}
  \Bigr\}.
\end{align}

After determining \(\rho\), each client sparsifies its coefficient vector 
\(\mathbf{c} = [c_0, \dots, c_{g(t)+o-1}]^{\top} \in \mathbb{R}^{(g(t)+o)\times1}\) 
for each activation edge by retaining only the top \(\rho\,g(t)\) coefficients in magnitude.  
Formally, this operation can be expressed as
\begin{equation}\label{eq3:ctop}
  \mathbf{c}_{\mathrm{top}}
  = \underset{\|\mathbf{c}'\|_{0} = \rho\,g(t)}{\arg\min}
    \bigl\|\mathbf{c} - \mathbf{c}'\bigr\|_{2},
\end{equation}
where \(\|\cdot\|_{2}\) denotes the $\ell_2$ norm.  
Finally, each client uploads \(\mathbf{c}_{\mathrm{top}}\) along with the remaining model parameters.
The overall process of CG-FKAN is summarized in \autoref{alg:proposed}.

\begin{algorithm}
\caption{Proposed CG-FKAN}
\label{alg:proposed}
\begin{algorithmic}[1]
\State Initialize parameters $w_g^{(-1)}$, $g_0$, $T_p$ and $\{\delta_j\}_{j=1}^{\lfloor T/T_p\rfloor}$
\For{global round $t = 0, 1, \ldots, T$}
    \State Server randomly selects a subset of clients $\mathcal{S}_{t}$ 
    \State Server broadcasts $(w_g^{(t-1)},g(t))$ to all clients in $\mathcal{S}_t$
    \For{client $k \in S_{t}$ \textbf{in parallel}}
        \State Initialize local model: $w_k^{(t-1)} \gets w_g^{(t-1)}$
            \State Compute KAN $h_k({x}_k)$ and loss $f_k(w_k^{(t-1)})$
            \State Update local model:
            \State $\quad w_k^{(t)} \gets w_k^{(t-1)} - \eta \nabla f_k(w_k^{(t-1)})$
            \If{$R_u(t)>R_{\rm budget}$}
                \State Compute ratio $\rho$ satisfying \eqref{eq2:ratio}
                \State Sparsify $\mathbf{c}_{\mathrm{top}}$ by computing \eqref{eq3:ctop}
            \EndIf
            \State Send the model parameters to the server            
    \EndFor
    \State Server aggregates parameters:
    \State $\quad w_g^{(t)} \gets \frac1{|\mathcal{S}_t|}\sum_{k\in\mathcal{S}_t}w_k^{(t)}$
\EndFor
\State \Return $w_g^{(T)}$
\end{algorithmic}
\end{algorithm}

\subsection{Complexity and Error Analysis}
We compare the computational complexity of CG-FKAN with that of the optimal sparsification method.  
For a fair comparison, the optimal sparsification under a given compression ratio \(\rho\) is formulated as:
\begin{align}\label{eq4:opt}
\mathbf{c}_{\mathrm{opt}}
= \underset{\|\mathbf{c}'\|_0 = \rho\,g(t)}{\arg\min}
\bigl\|\Phi\,\mathbf{c} - \Phi\,\mathbf{c}'\bigr\|_2,    
\end{align}
where \(\Phi = [B_0(x), \dots, B_{g(t)+o-1}(x)] \in \mathbb{R}^{1 \times (g(t)+o)}\) denotes the B-spline basis vector.  
Solving~\eqref{eq4:opt} requires exhaustive search since all possible combinations of \(\rho\,g(t)\) coefficients must be evaluated from \(g(t){+}o\) candidates.  
Therefore, the optimal method entails a computational complexity of
\begin{align}
    \mathcal{O}\Biggl(\binom{g(t)+o}{\rho\,g(t)}\Biggr),
\end{align}
which increases combinatorially with \(g(t)\).  
In contrast, CG-FKAN only requires sorting a coefficient vector of length \(g(t)+o\), resulting in a complexity as:
\begin{align}
    \mathcal{O}\bigl((g(t)+o)\,\log (g(t)+o)\bigr).
\end{align}
Therefore, as \(g(t)\) increases, the gap between the optimal and CG-FKAN widens rapidly. 
In addition, we derive an explicit error bound between the CG-FKAN and the optimal sparsification to evaluate the performance trade-off in B-spline approximation.

\noindent\textbf{Proposition.}
Let the B-spline basis function vector be
\[
\Phi(x) = [B_0(x),\,B_1(x),\dots,\,B_{g(t)+o-1}(x)]^{\!\top}
\]
and
\[
\mathbf{c}_{\mathrm{top}}
=\underset{\|\mathbf{c}'\|_0=m}{\arg\min}\|\mathbf{c}-\mathbf{c}'\|_2,
\quad
\mathbf{c}_{\mathrm{opt}}
=\underset{\|\mathbf{c}'\|_0=m}{\arg\min}\|\Phi(x)\,\bigl(\mathbf{c}-\mathbf{c}'\bigr)\|_2.
\]

The approximation errors are given by:
\[
e_{\mathrm{top}}=\|\Phi(x)\,\{\mathbf{c}-\mathbf{c}_{\mathrm{top}}\}\|_2,
\quad
e_{\mathrm{opt}}=\|\Phi(x)\,\{\mathbf{c}-\mathbf{c}_{\mathrm{opt}}\}\|_2.
\]
Then, the following holds:
\[
  e_{\mathrm{top}} \;<\; o\cdot\,2^o\;e_{\mathrm{opt}}.
\]
\noindent\textit{Proof.}
Let \(\mathbf{d}_{\mathrm{top}}=\mathbf{c}-\mathbf{c}_{\mathrm{top}}\) and \(\mathbf{d}_{\mathrm{opt}}=\mathbf{c}-\mathbf{c}_{\mathrm{opt}}\).
\begin{align*}
e_{\mathrm{top}}
&= \|\Phi\,\mathbf{d}_{\mathrm{top}}\|_{2}\\
&\underset{(a)}{\le} \sup_{\|\mathbf{u}\|_2=1}\|\Phi\,\mathbf{u}\|_2\,\|\mathbf{d}_{\mathrm{top}}\|_2\\
&\underset{(b)}{\le} \sup_{\|\mathbf{u}\|_2=1}\|\Phi\,\mathbf{u}\|_2\,\|\mathbf{d}_{\mathrm{opt}}\|_2\\
&= \frac{\sup_{\|\mathbf{u}\|_2=1}\|\Phi\,\mathbf{u}\|_2}{\inf_{\|\mathbf{u}\|_2=1}\|\Phi\,\mathbf{u}\|_2\;}
    \bigl(\inf_{\|\mathbf{u}\|_2=1}\|\Phi\,\mathbf{u}\|_2\;\|\mathbf{d}_{\mathrm{opt}}\|_2\bigr)\\
&\underset{(c)}{\le} \frac{\sup_{\|\mathbf{u}\|_2=1}\|\Phi\,\mathbf{u}\|_2}{\inf_{\|\mathbf{u}\|_2=1}\|\Phi\,\mathbf{u}\|_2\;} e_{\mathrm{opt}} \;\underset{(d)}{<}\; o\cdot\,\,2^o\;e_{\mathrm{opt}}. 
\end{align*}
Here, $(a)$ and $(c)$ follow from the operator norm inequality\(\|\Phi \mathbf{d}\|_2 \le \sup_{\|\mathbf{u}\|_2=1} \|\Phi \mathbf{u}\|_2 \cdot \|\mathbf{d}\|_2\) and its corresponding lower bound,
$(b)$ uses the fact that \(\|\mathbf{d}^{\mathrm{top}}\|_2 \le \|\mathbf{d}^{\mathrm{opt}}\|_2\), since \(\mathbf{c}^{\mathrm{top}}\) minimizes the Euclidean distance to~\(\mathbf{c}\),
and $(d)$ applies the known upper bound on the condition number of the B-spline basis function, \(\  {\sup_{\|\mathbf{u}\|_2=1}\|\Phi\,\mathbf{u}\|_2}/{\inf_{\|\mathbf{u}\|_2=1}\|\Phi\,\mathbf{u}\|_2\;}< o\cdot\,2^o\)~\cite{scherer1999new}.\qed

Rather than raising concerns over the exponential term $o\cdot\,2^o$, we observe that as the optimal error $e_{\text{opt}}$ approaches zero, the proposed approximation error $e_{\text{top}}$ consequently vanishes asymptotically. 

\section{Experiment and Results}\label{ch4:experiments}
\subsection{Experimental Setup}
We evaluate the approximation capability of the proposed method through regression tasks on benchmark functions where KAN shows strong performance~\cite{liu2024kan1}, as summarized in Table~\ref{tab:functions}.  
The network configurations for Feynman eq.~I.30.3, Feynman eq.~I.37.4, Bessel, and Legendre functions are set to \([n_0,n_1,n_2,n_3]=[3,5,5,1]\), \([3,3,2,1]\), \([2,2,2,1]\), and \([2,2,1]\), respectively.  
To construct the FL environment, we consider both IID and non-IID scenarios.  
The non-IID data distributions are generated using a Dirichlet allocation with \(\alpha \in \{100.0, 10.0, 1.0\}\).  
We set the total number of clients to \(N=100\), the total communication rounds to \(T=1000\), and the number of local epochs to \(5\).  
In each round, \(10\%\) of clients are randomly selected to participate in training.

\begin{table}[t]
\caption{Analytical functions used in regression tasks.}
\label{tab:functions}
\centering
\begin{tabular}{cc}
\toprule
Function & Expression\\
\midrule
Feynman eq. I.30.3 & $f(I_0,n,\theta) = I_0 \cdot {\sin^2\left( {n\theta}/{2} \right)}/{\sin^2\left({\theta}/{2} \right)}$ \\
Feynman eq. I.37.4 & $f(I_1,I_2,\delta) = I_1 + I_2 + 2\sqrt{I_1 I_2} \cos(\delta)$ \\
Bessel function & $f(\nu, x) = \sum_{m=0}^{\infty} \frac{(-1)^m}{m! \, \Gamma(m + \nu + 1)} \left( \frac{x}{2} \right)^{2m + \nu}$ \\
Legendre polynomial & $f(n, z) = \frac{1}{2^n n!} \cdot \frac{d^n}{dz^n} (z^2 - 1)^n
$\\
\bottomrule
\end{tabular}
\end{table}

The grid-extended parameters are set to \(T_P = 200\), \(g_0 = 3\), and \(\{\delta_j\}_{j=1}^{\lfloor T/T_P \rfloor} = \{2, 7, 27, 47\}\).  
Since the number of parameters in a KAN model depends on the grid size, the communication budget \(R_{\text{budget}}\) is fixed to match the parameter size of a KAN with grid size 10, i.e.,  
\begin{align}
R_{\text{budget}} = b\left(C_0 + \sum_{i=0}^{L-1} 10\,\omega_i\right).
\end{align}
For fair comparison, we evaluate three baselines: (i) fixed-grid KAN (ii) grid-extended KAN, and (iii) MLPs whose parameter count is matched to that of a KAN with grid size 100.  
Additional hyperparameter details are available in our open-source repository.\footnote{\url{https://github.com/bgbhbf/CG-FKAN}}

\subsection{Results}
\subsubsection{Impact of CG-FKAN}
\begin{figure}
    \includegraphics{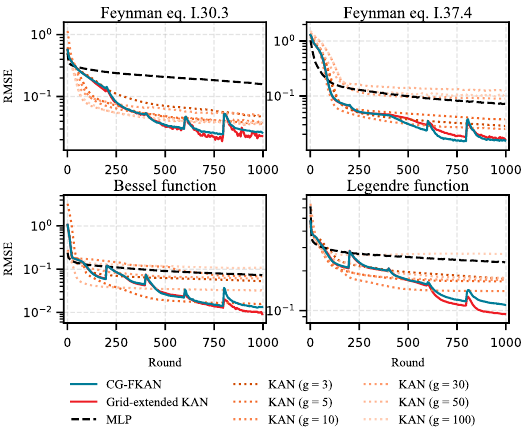} 
    \caption{RMSE results of CG-FKAN compared with baselines across four benchmark functions (Feynman eq. I.30.3, Feynman eq. I.37.4, Bessel, and Legendre functions). 
    }
    \label{fig:RMSE}
\end{figure}

To evaluate the effectiveness of CG-FKAN, we conduct experiments under the test environment illustrated in \cref{fig:RMSE}.  
In this setting, we compare FL adopting grid-extended KAN, which serves as an upper bound since all parameters are transmitted without communication constraints.  
The proposed CG-FKAN consistently outperforms all baselines except for grid-extended KAN.  

In particular, for Feynman Eq.~I.37.4, the RMSE of CG-FKAN ($1.534\times10^{-2}$) is slightly lower than that of grid-extended KAN ($1.620\times10^{-2}$), corresponding to a reduction of approximately $5.3\%$.  
This improvement arises from the pruning effect induced by coefficient sparsification.  
In the Bessel function case, however, CG-FKAN shows a larger gap, where its RMSE ($1.334\times10^{-2}$) is about $31.6\%$ higher than that of grid-extended KAN ($9.123\times10^{-3}$).  
Nevertheless, CG-FKAN clearly surpasses the best fixed-grid KAN ($1.543\times10^{-2}$), achieving a reduction of approximately $13.6\%$.  
Overall, despite the performance gap from grid-extended KAN, CG-FKAN remains superior to all fixed-grid baselines.

\subsubsection{Impact of Data Heterogeneity}
We further evaluate CG-FKAN under data-skewed environments to assess its robustness against heterogeneity.  
As summarized in \cref{tab:data_skew}, CG-FKAN consistently achieves performance comparable to grid-extended KAN.  
For reference, the best fixed-grid KAN is selected based on the lowest RMSE among grid sizes \(\{3, 5, 10, 30, 50, 100\}\).  
These best cases typically occur at smaller grids, where communication cost is low but model expressiveness is limited.  
Consequently, even the best fixed-grid KAN exhibits higher RMSE than CG-FKAN, which dynamically utilizes its communication budget \(R_{\text{budget}}\) to retain the most informative parameters.  

For instance, in the Bessel function task with \(\alpha = 1.0\), CG-FKAN attains an RMSE of \(1.199\times10^{-2}\), outperforming the best fixed-grid KAN (\(1.508\times10^{-2}\) at \(g=5\)) and approaching the grid-extended KAN (\(9.409\times10^{-3}\)).  
These results confirm that CG-FKAN effectively balances communication efficiency and approximation accuracy under data heterogeneity.

\begin{table}[!t]
\centering
\caption{RMSE results under different Dirichlet distributions. 
For fixed-grid KAN, only the best grid is reported.}
\label{tab:data_skew}
\scriptsize 
\resizebox{\columnwidth}{!}{
\begin{tabular}{cccc}
\toprule
\textbf{Function} ($\alpha$)& \textbf{Grid-extended KAN} & \textbf{CG-FKAN} & \textbf{Best fixed grid KAN} \\
\midrule
I.30.3 (1.0) & {$2.329\times10^{-2}$} & {$2.525\times10^{-2}$} & $3.634\times10^{-2}$ (g=10)  \\
I.37.4 (1.0) & {$1.458\times10^{-2}$} & { $1.400\times10^{-2}$} & $2.400\times10^{-2}$ (g=3) \\
Bessel (1.0) & {$9.409\times10^{-3}$} & { $1.199\times10^{-2}$} & $1.508\times10^{-2}$ (g=5) \\
Legendre (1.0) & { $9.899\times10^{-2}$} & { $9.276\times10^{-2}$} & $1.402\times10^{-1}$ (g=10) \\
\midrule
I.30.3 (10.0) & { $2.195\times10^{-2}$} & { $2.274\times10^{-2}$} & $3.564\times10^{-2}$ (g=10) \\
I.37.4 (10.0) & { $1.105\times10^{-2}$} & { $1.306\times10^{-2}$} & $2.718\times10^{-2}$ (g=10) \\
Bessel (10.0) & { $8.316\times10^{-3}$} & { $1.216\times10^{-2}$} & $1.542\times10^{-2}$ (g=5) \\
Legendre (10.0) & { $8.856\times10^{-2}$} & { $8.460\times10^{-2}$} & $1.404\times10^{-1}$ (g=10) \\
\midrule
I.30.3 (100.0) & { $2.219\times10^{-2}$} & { $2.322\times10^{-2}$} & $3.328\times10^{-2}$ (g=10) \\
I.37.4 (100.0) & { $1.076\times10^{-2}$} & { $1.299\times10^{-2}$} & $2.540\times10^{-2}$ (g=10) \\
Bessel (100.0) & { $7.908\times10^{-3}$} & { $1.166\times10^{-2}$} & $1.542\times10^{-2}$ (g=5) \\
Legendre (100.0) & { $8.690\times10^{-2}$} & { $9.886\times10^{-2}$} & $1.405\times10^{-1}$ (g=10) \\
\bottomrule
\end{tabular}
}
\end{table}

\subsubsection{Impact of Communication Efficiency}
\begin{figure}[t]
    \includegraphics{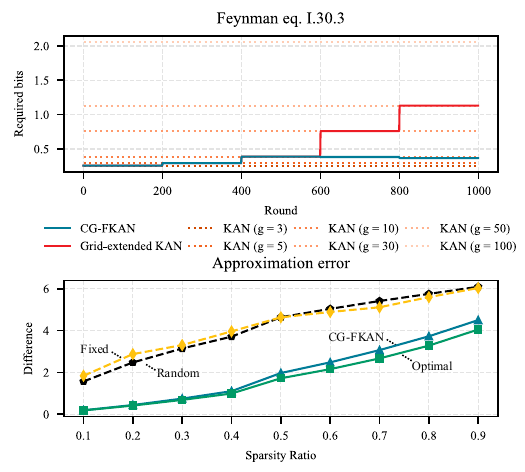} 
    \caption{Upper: Required bits per round to transmit parameters to the server during regression tasks on the Feynman equation I.30.3. 
    Lower: Approximation error versus sparsity ratio for the case $g=10$ and $o=3$, comparing different sparsification methods.}
    \label{fig:communication_cost}
\end{figure}

To analyze the trade-off between communication cost and approximation accuracy, we evaluate the number of bits required for model transmission and the corresponding error across different sparsification strategies, as shown in \cref{fig:communication_cost}.  
With 32-bit parameters (\(b=32\)), CG-FKAN requires fewer bits to satisfy the constrained rate \(R_{\text{budget}}\), saving approximately 2{,}170 bits compared to the fixed-grid KAN with \(g=10\) and 76{,}410 bits compared to the grid-extended KAN over the final 200 rounds in the upper panel.  
The lower panel shows that CG-FKAN achieves significantly lower error than random and fixed sparsification, while remaining close to the optimal case.  
Thus, CG-FKAN achieves a balanced trade-off between communication efficiency and accuracy, approaching grid-extended KAN while surpassing alternative sparsification methods.

\section{Conclusion}\label{ch5:conclusion}
In this letter, we proposed an algorithm to mitigate the communication burden inherent in FL-KAN, an aspect largely overlooked in prior studies on grid extension. 
The proposed CG-FKAN applies coefficient sparsification to preserve only the most informative spline components under a communication budget.
Furthermore, we derive an upper bound on the approximation error of CG-FKAN with respect to optimal sparsification. 
Numerical results demonstrate that CG-FKAN consistently outperforms the best fixed-grid KAN, while achieving performance comparable to grid-extended KAN with significantly reduced communication cost.

\bibliographystyle{IEEEtran}
\bibliography{reference}

\begin{thebibliography}{10}
\providecommand{\url}[1]{#1}
\csname url@samestyle\endcsname
\providecommand{\newblock}{\relax}
\providecommand{\bibinfo}[2]{#2}
\providecommand{\BIBentrySTDinterwordspacing}{\spaceskip=0pt\relax}
\providecommand{\BIBentryALTinterwordstretchfactor}{4}
\providecommand{\BIBentryALTinterwordspacing}{\spaceskip=\fontdimen2\font plus
\BIBentryALTinterwordstretchfactor\fontdimen3\font minus \fontdimen4\font\relax}
\providecommand{\BIBforeignlanguage}[2]{{%
\expandafter\ifx\csname l@#1\endcsname\relax
\typeout{** WARNING: IEEEtran.bst: No hyphenation pattern has been}%
\typeout{** loaded for the language `#1'. Using the pattern for}%
\typeout{** the default language instead.}%
\else
\language=\csname l@#1\endcsname
\fi
#2}}
\providecommand{\BIBdecl}{\relax}
\BIBdecl

\bibitem{zhu2020toward}
G.~Zhu, D.~Liu, Y.~Du, C.~You, J.~Zhang, and K.~Huang, ``Toward an intelligent edge: Wireless communication meets machine learning,'' \emph{IEEE Commun. Mag.}, vol.~58, no.~1, pp. 19--25, 2020.

\bibitem{kairouz2021advances}
P.~Kairouz and H.~McMahan, \emph{Advances and Open Problems in Federated Learning}, ser. Found. Trends Mach. Learn.\hskip 1em plus 0.5em minus 0.4em\relax Now Publishers, 2021.

\bibitem{mcmahan2017communication}
B.~McMahan, E.~Moore, D.~Ramage, S.~Hampson, and B.~A. y~Arcas, ``Communication-efficient learning of deep networks from decentralized data,'' in \emph{Proc. 20th Int. Conf. Artif. Intell. Stat.}, Apr. 2017, pp. 1273--1282.

\bibitem{yang2021federated}
Z.~Yang, M.~Chen, K.-K. Wong, H.~V. Poor, and S.~Cui, ``Federated learning for {6G}: Applications, challenges, and opportunities,'' \emph{Engineering}, vol.~8, pp. 33--41, 2022.

\bibitem{9415623}
D.~C. Nguyen, M.~Ding, P.~N. Pathirana, A.~Seneviratne, J.~Li, and H.~Vincent~Poor, ``Federated learning for internet of things: A comprehensive survey,'' \emph{IEEE Commun. Surveys Tuts.}, vol.~23, no.~3, pp. 1622--1658, Apr. 2021.

\bibitem{khan2021federated}
L.~U. Khan, W.~Saad, Z.~Han, E.~Hossain, and C.~S. Hong, ``Federated learning for internet of things: Recent advances, taxonomy, and open challenges,'' \emph{IEEE Commun. Surv. Tutor.}, vol.~23, no.~3, pp. 1759--1799, 2021.

\bibitem{nascita2024survey}
A.~Nascita, G.~Aceto, D.~Ciuonzo, A.~Montieri, V.~Persico, and A.~Pescap{\'e}, ``A survey on explainable artificial intelligence for internet traffic classification and prediction, and intrusion detection,'' \emph{IEEE Commun. Surveys Tuts.}, vol.~27, no.~5, pp. 3165--3198, Oct. 2025.

\bibitem{jagatheesaperumal2024enabling}
S.~K. Jagatheesaperumal, M.~Rahouti, A.~Alfatemi, N.~Ghani, V.~K. Quy, and A.~Chehri, ``Enabling trustworthy federated learning in industrial {IoT}: bridging the gap between interpretability and robustness,'' \emph{IEEE Internet Things Mag.}, vol.~7, no.~5, pp. 38--44, Sep. 2024.

\bibitem{Zeleke2024}
S.~N. Zeleke and M.~Bochicchio, ``Federated kolmogorov-arnold networks for health data analysis: A study using ecg signal,'' in \emph{Proc. IEEE Int. Conf. Big Data}, Washington, DC, United States, Dec. 2024.

\bibitem{lee2025unifiedbenchmarkfederatedlearning}
Y.~Lee, J.~Gong, and J.~Kang, ``A unified benchmark of federated learning with kolmogorov-arnold networks for medical imaging,'' \emph{arXiv preprint arXiv:2504.19636}, 2025.

\bibitem{Zeydan2025}
E.~Zeydan, C.~J. Vaca-Rubio, L.~Blanco, R.~Pereira, M.~Caus, and A.~Aydeger, ``{F\textnormal{-}KANs}: Federated kolmogorov-arnold networks,'' in \emph{Proc. IEEE 22nd Consum. Commun. Netw. Conf.}, Las Vegas, NV, United States, Jan. 2025.

\bibitem{11106287}
F.~Li, X.~Chen, K.-Y. Lam, L.~Wang, and X.~Liu, ``A secure wireless traffic prediction with federated learning,'' \emph{IEEE Trans. Veh. Technol.}, pp. 1--12, Jul. 2025 (Early Access).

\bibitem{liu2024kan1}
Z.~Liu, Y.~Wang, S.~Vaidya, F.~Ruehle, J.~Halverson, M.~Solja{\v{c}}i{\'c}, T.~Y. Hou, and M.~Tegmark, ``{KAN}: Kolmogorov-arnold networks,'' in \emph{Proc. Int. Conf. Learn. Represent.}, Singapore, Apr. 2025.

\bibitem{Gong2023}
J.~Gong, O.~Simeone, and J.~Kang, ``Compressed particle‑based federated bayesian learning and unlearning,'' \emph{IEEE Commun. Lett.}, vol.~27, no.~2, pp. 556--560, 2023.

\bibitem{Yi_kong2024}
Y.~Kong, W.~Yu, S.~Xu, F.~Yu, Y.~Xu, and Y.~Huang, ``Two birds with one stone: Toward communication and computation efficient federated learning,'' \emph{IEEE Commun. Lett.}, vol.~28, no.~9, pp. 2106--2110, 2024.

\bibitem{li2022federated}
Q.~Li, Y.~Diao, Q.~Chen, and B.~He, ``Federated learning on non-iid data silos: An experimental study,'' in \emph{Proc. IEEE International Conference on Data Engineering (ICDE)}, Kuala Lumpur, Malaysia, May 2022.

\bibitem{Sattler2019SparseBinaryCompression}
F.~Sattler, S.~Wiedemann, K.-R. M{\"u}ller, and W.~Samek, ``Sparse binary compression: Towards distributed deep learning with minimal communication,'' in \emph{Proc. Int. Joint Conf. Neural Netw.}, Jul. 2019, pp. 1--8.

\bibitem{scherer1999new}
K.~Scherer and A.~Y. Shadrin, ``New upper bound for the {B}‐spline basis condition number,'' \emph{J. Approx. Theory}, vol.~99, no.~2, pp. 217--229, 1999.

\end{thebibliography}

\end{document}